\documentclass[11pt]{article}

\usepackage{multirow}
\usepackage{mathtools}
\usepackage{booktabs}
\usepackage{amsmath}
\usepackage{algorithm}
\usepackage{algorithmic}

\usepackage{enumitem}

\usepackage{graphicx} 
\usepackage{caption}

\usepackage{booktabs}
\usepackage{multirow}
\usepackage[table,xcdraw]{xcolor}
\usepackage{tabularx} 
\usepackage{amsmath}

\usepackage{comment}

\usepackage{amssymb}
\usepackage[table]{xcolor}
\usepackage{booktabs}
\definecolor{tableheader}{RGB}{41,82,122}
\definecolor{tablesection}{RGB}{214,234,248}

\definecolor{mygray}{gray}{0.92}

\newcolumntype{Y}{>{\centering\arraybackslash}X}

\usepackage[final]{acl}

\usepackage{times}
\usepackage{latexsym}

\usepackage[T1]{fontenc}

\usepackage[utf8]{inputenc}

\usepackage{microtype}

\usepackage{inconsolata}

\usepackage{graphicx}

\title{CRAFTQA: A Code‑Driven Adaptive Framework for Complex Structured Data Reasoning}

\author{
  \textbf{Chengtao Gan}\textsuperscript{$\spadesuit$},
  \textbf{Zhiqiang Liu}\textsuperscript{$\spadesuit$},
  \textbf{Long Jin}\textsuperscript{$\spadesuit$},
  \textbf{Yushan Zhu}\textsuperscript{$\heartsuit$},
  \\
  \textbf{Lei Liang}\textsuperscript{$\clubsuit\diamondsuit$},
  \textbf{Wen Zhang}\textsuperscript{$\spadesuit$$\diamondsuit$\textdagger}
  \\
  $^{\spadesuit}$Zhejiang University 
  $^{\clubsuit}$Ant Group \\
  $^{\heartsuit}$JIUTIAN Research, Beijing, China \\
  $^{\diamondsuit}$ZJU-Ant Group Joint Lab of Knowledge Graph \\
  \texttt{\{chengtaogan,zhang.wen\}@zju.edu.cn}
}

\begin{document}
\maketitle

\begin{abstract}
Real-world scenarios involve massive heterogeneous structured data (e.g., tables, knowledge graphs), making effective reasoning over such diverse data increasingly important. Unified structured data question answering has emerged as a prominent research trend, aiming to answer natural language questions across different structured data types within a single framework. However, existing unified methods share a common limitation: they rely on a set of predefined functions, which restricts their ability to perform complex reasoning beyond these predefined operations. To overcome this fundamental limitation, we propose \textbf{CRAFTQA}, a novel adaptive code-driven framework comprising two core modules, CodeSTEP and CRAFT. 
The \textbf{CodeSTEP} module is a paradigm that generates a complete executable Python code sequence, which contains step-by-step code-based reasoning operations based on the question.
The \textbf{CRAFT} module dynamically generates custom code functions for operations beyond the predefined function set, and seamlessly integrates with CodeSTEP to significantly enhance flexibility in handling complex reasoning. Comprehensive experiments on multiple structured datasets demonstrate that CRAFTQA achieves remarkable improvements in complex reasoning scenarios compared to existing unified methods.
\end{abstract}

\section{Introduction}

\begingroup
  \renewcommand{\thefootnote}{\fnsymbol{footnote}}
  \setcounter{footnote}{0}
  \footnotetext{\textsuperscript{\textdagger} Corresponding author}
\endgroup

Structured data (e.g., tables, knowledge graphs) organize information in well-defined formats, enabling efficient storage, retrieval, and computation \cite{tan2024struct}. Real-world scenarios involve massive amounts of structured data. In the era of Large Language Models (LLMs), effectively \textbf{leveraging structured data sources} is crucial for enhancing modern AI systems, particularly in improving factual accuracy, reducing hallucinations, and supporting complex reasoning capabilities \cite{yang2024give}.

Natural language reasoning over structured data has become increasingly important \cite{yu2024natural}. Data-specific methods have been developed for particular data structures like tables or knowledge graphs (e.g., TAPEX \cite{liu2021tapex} and DecAF \cite{yu2022decaf}). However, real-world scenarios often require reasoning across different types of data sources \cite{yin2020tabert}. Given the massive heterogeneous structured data in real-world applications, effectively reasoning over such diverse data has become increasingly critical. This has driven significant research interest in \textbf{unified structured data question answering methods}, which aim to handle multiple structured data types within a single framework \cite{zhang2025trustuqa}.
For example, retrieval-based unified methods like StructGPT \cite{jiang2023structgpt} and Readi \cite{cheng2024call} access raw data through predefined functions. To further enhance trustworthiness, TrustUQA \cite{zhang2025trustuqa} obtains answers through a unified query language without inputting extensive raw data into the LLM.

However, existing \textbf{unified methods share a common limitation: they rely on a set of predefined functions} \cite{jiang2023structgpt,cheng2024call,zhang2025trustuqa}. Complex tasks involving sophisticated computation and advanced logical reasoning often require operations beyond these predefined functions \cite{gao2023pal}, which fundamentally \textbf{limits their ability to handle complex reasoning beyond predefined operations}.

Methods such as PoT \cite{chen2022program} and PAL \cite{gao2023pal} have demonstrated that LLMs can better solve complex numerical reasoning tasks by generating Python programs. We believe that decoupling reasoning from computation through code-based programs can effectively enhance LLMs' reasoning capabilities while ensuring transparency and interpretability.

Inspired by code-based program reasoning \cite{chen2022program}, and to \textbf{address the fundamental limitation of existing unified methods}, we propose \textbf{CRAFTQA}, a code-driven adaptive framework for unified structured data question answering, comprising two core modules: \textbf{Code}-Based \textbf{S}tepwise \textbf{T}ransparent \textbf{E}xecution \textbf{P}aradigm (\textbf{CodeSTEP}) and \textbf{C}ode-Based \textbf{R}easoning for \textbf{A}daptive \textbf{F}unction \textbf{T}ailoring (\textbf{CRAFT}).

\textbf{CodeSTEP} is a custom code paradigm that generates \textbf{complete executable Python code} sequences for structured data reasoning, effectively separating reasoning from computation. This code-based approach aligns better with LLMs' inherent reasoning patterns \cite{gao2023pal,chen2022program}, offering enhanced reasoning capabilities, flexible representations, and transparent processing workflows. Moreover, CodeSTEP's explicit reasoning steps serve as crucial context for the CRAFT module to understand complex scenarios.

\textbf{CRAFT} is designed to dynamically handle scenarios beyond predefined functions. We proposed CRAFT to \textbf{overcome the fundamental limitation of existing unified structured data QA methods} that can only operate within predefined functions.
\textbf{CRAFT can generate dedicated code for specific reasoning steps and seamlessly integrate with the main CodeSTEP execution.} This design enhances framework flexibility while maintaining verifiability, enabling the handling of ``out-of-predefined'' operations that existing methods cannot address, thereby making CRAFTQA particularly suitable for complex reasoning tasks.

We conducted comprehensive experiments on 6 datasets across 3 structured data types: Table (WikiSQL \cite{zhong2017seq2sql} and TableBench \cite{wu2025tablebench}), Knowledge Graph (WebQSP \cite{yih2016value}), and Temporal Knowledge Graph (CronQuestions \cite{saxena2021question}). To validate CRAFT's effectiveness in ``out-of-predefined'' scenarios, we constructed WikiSQL-E by extracting QA pairs from WikiSQL that potentially require ``out-of-predefined'' operations. Results show that CRAFTQA achieves \textbf{state-of-the-art} performance on \textbf{complex reasoning tasks} while maintaining strong standard reasoning capabilities. Furthermore, CRAFTQA demonstrates generalizability across diverse backbone LLM families, where CRAFTQA with smaller open-source backbone LLMs even outperforms advanced baselines with larger closed-source models.

In summary, contributions of this paper are:
\begin{itemize}
\item[$\bullet$] We present \textbf{CRAFTQA}, a flexible code-driven framework for unified structured data question answering, comprising \textbf{CodeSTEP} for step-by-step code-based reasoning and \textbf{CRAFT} for dynamic function customization to handle ``out-of-predefined'' operations.
\item[$\bullet$] To the best of our knowledge, we are the \textbf{first to implement unified adaptive structured data reasoning in a custom code-based form}, enabling complex reasoning tasks that existing methods struggle to address.
\item[$\bullet$] Experiments on 6 datasets across 3 structured data types demonstrate that CRAFTQA significantly \textbf{outperforms existing unified methods} in complex reasoning scenarios while maintaining strong standard reasoning performance, and exhibits robust generalizability across diverse backbone LLM families.
\end{itemize}

\section{Related Work}

\subsection{Structured Data Question Answering}
Structured Data QA plays an increasingly important role in human-computer interaction across healthcare \cite{yang2024kg,huang2021knowledge}, finance \cite{liu2023tab}, and information retrieval \cite{zhang2022improving}. Research in this field has evolved along two primary directions. Single \textbf{data-type specific} methods focus on reasoning over a specific data structure, such as tables \cite{zha2023tablegpt}, KGs \cite{song2023advancements}, or TKGs \cite{QC_MHM_xue2024question}. Recent advancements include ReasoningLM \cite{jiang2023reasoninglm}, which utilizes a code-generation-based paradigm to improve multi-hop reasoning. In contrast, \textbf{unified} methods aim to support reasoning across multiple structured data types simultaneously \cite{khashabi2020unifiedqa}, such as StructGPT \cite{jiang2023structgpt}, which handles KGs, tables, and databases.

\subsection{LLM-Based Unified Frameworks}
With the rapid advancement of large language models, several unified frameworks have been proposed for structured data question answering. StructGPT \cite{jiang2023structgpt} is an iterative reading-then-reasoning framework that generates answers based on collected evidence. Readi \cite{cheng2024call} is a reasoning-path-editing framework that collects KG evidence through edited reasoning paths. TrustUQA \cite{zhang2025trustuqa} employs Condition Graph and a two-layer query approach to uniformly support tables, KGs, and TKGs.

\subsection{Code-Based Reasoning}
Recent research has demonstrated that code-based methods can enhance reasoning capabilities in LLMs \cite{yang2025code}. Program of Thought (PoT) \cite{chen2022program} and PAL \cite{gao2023pal} show that executable code can decompose complex problems into manageable computational steps, facilitating more interpretable and controllable reasoning processes.

\section{Methodology}

\begin{figure*}[t]
\centering
\includegraphics[width=0.9\textwidth]{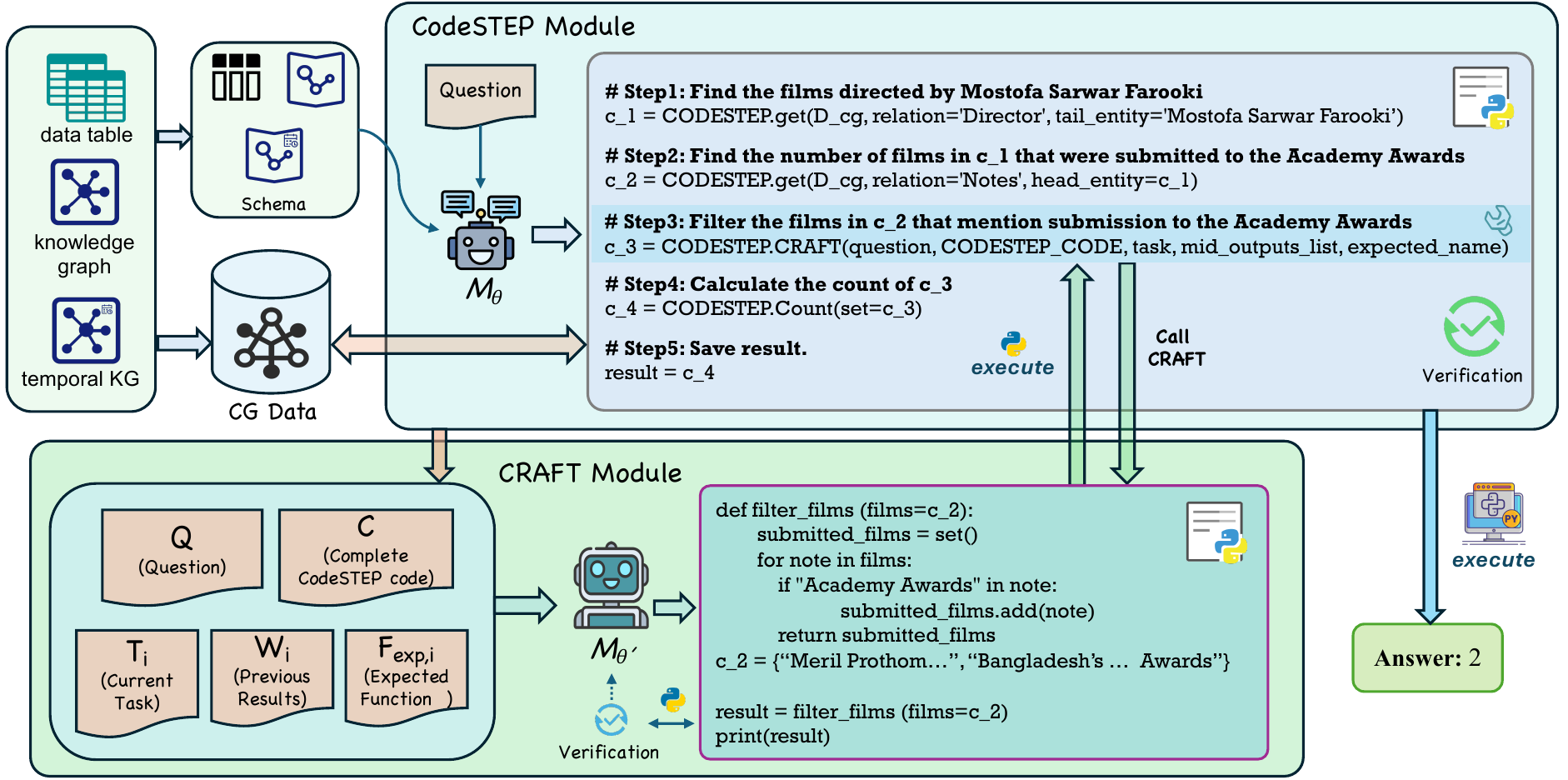}

\caption{Overview of CRAFTQA framework. The framework comprises two synergistic modules: CodeSTEP for stepwise code-based reasoning and CRAFT for dynamic custom function generation.}
\vspace{-\baselineskip}
\label{fig:model_arch}
\end{figure*}

\subsection{Overview}

CRAFTQA is a unified code-driven framework for structured data question answering. As illustrated in Figure \ref{fig:model_arch}, it comprises two synergistic modules: \textbf{CodeSTEP} (\textbf{Code}-based \textbf{S}tepwise \textbf{T}ransparent \textbf{E}xecution \textbf{P}aradigm) for stepwise code-based reasoning, and \textbf{CRAFT} (\textbf{C}ode-based \textbf{R}easoning for \textbf{A}daptive \textbf{F}unction \textbf{T}ailoring) for dynamic custom function generation.

Given a data source $\mathcal{D}$ and question $\mathcal{Q}$, we transform $\mathcal{D}$ into a data schema $\mathcal{D}_{schema}$ containing structural metadata, and a Condition Graph $\mathcal{D}_{cg}$ \cite{zhang2025trustuqa} for unified querying. An LLM $\mathcal{M}_\theta$ with few-shot prompt $\mathcal{P}$ generates executable code $\mathcal{C} = \{c_i\}_{i=1}^{n}$, executed on $\mathcal{D}_{cg}$ to yield the answer $\mathcal{A}$:
\begin{equation}
\label{eq:code_gen}
\mathcal{M}_\theta(\mathcal{D}_{schema}, \mathcal{Q}, \mathcal{P}) \rightarrow \mathcal{C},
\end{equation}
\begin{equation}
\label{eq:execute}
\textsc{Exec}(\mathcal{C}, \mathcal{D}_{cg}) \rightarrow \mathcal{A}.
\end{equation}

\subsection{CodeSTEP Module}

\paragraph{Stepwise Code Generation.}
CodeSTEP decomposes code generation into two phases:
\noindent\textbf{(1) Query Analysis.} Given question $\mathcal{Q}$ and data schema $\mathcal{D}_{schema}$, $\mathcal{M}_\theta$ constructs a reasoning path $\mathcal{S} = \{s_i\}_{i=1}^{n}$, where each $s_i$ is a natural language reasoning step.
\noindent\textbf{(2) Stepwise Code Construction.} For each $s_i$, $\mathcal{M}_\theta$ generates a corresponding code $c_i$, constructing the complete executable sequence $\mathcal{C} = \{c_i\}_{i=1}^{n}$, as formulated in Equation \ref{eq:code_gen}.

\paragraph{Condition Graph Query Operation.}
The primary data retrieval operation is the query function $get$, which retrieves target entities from $\mathcal{D}_{cg}$:
\begin{equation}
\label{eq:cg_query}
get: (\mathcal{D}_{cg}, R, E_h, E_t, \delta_t, K, V, \delta_v) \rightarrow O,
\end{equation}

where $R$ denotes the relation (column header for tables, edge type for KGs), $E_h$ and $E_t$ are head and tail entity sets, $\delta_t \in \{=, >, <, \geq, \leq\}$ is the comparison operator for $E_t$, and $O$ is the output entity set. $K$ specifies a conditional attribute (e.g., temporal property), $V$ provides the threshold value of $K$, and $\delta_v$ is the comparison operator of $V$.

\smallskip
\noindent\fbox{\parbox{0.95\columnwidth}{
\textbf{Example:} 

\centering
$get(E_h{=}\texttt{None}, R{=}\texttt{`Won'}, E_t{=}\texttt{`Nobel Prize'},$ \\
$K{=}\texttt{`Year'}, V{=}2000, \delta_v{=}\texttt{`>'}),$ \\[0.5ex]
\raggedright
retrieves all `Nobel Prize' winners where the \texttt{`Year'} attribute exceeds 2000.
}}
\smallskip

\paragraph{Semantic Entity Alignment.}
Entities in generated code may not exactly match those in $\mathcal{D}_{cg}$ due to lexical variations. We address this via Sentence-BERT \cite{reimers2019sentence} alignment. For an entity $e$ in the code and candidates $\mathcal{E} = \{e_j\}_{j=1}^{m}$ in $\mathcal{D}_{cg}$:
\begin{equation}
\label{eq:alignment}
e^* = \underset{e_j \in \mathcal{E}}{\arg\max} \ \cos\bigl(\phi(e), \phi(e_j)\bigr),
\end{equation}
where $\phi(\cdot)$ denotes Sentence-BERT embedding and $\cos(\cdot,\cdot)$ computes cosine similarity. This alignment is performed automatically within $get$, ensuring accurate entity matching without explicit handling in generated code.

\paragraph{Predefined Function Set.}
Beyond $get$, CodeSTEP provides set and algebraic operations, as detailed in Table \ref{tab:support-ops}. These collectively form the predefined function set:
\begin{equation}
\label{eq:fpred}
\mathcal{F}_{pred} = \{get\} \cup \mathcal{F}_{set} \cup \mathcal{F}_{cal},
\end{equation}
where $\mathcal{F}_{set} = \{f_{\cup}, f_{\cap}, f_{-}\}$ and $\mathcal{F}_{cal} = \{f_{\min}, f_{\max}, f_{avg}, f_{cnt}, f_{sum}\}$.

During the reasoning process, each code step $c_i$ may utilize functions from $\mathcal{F}_{pred}$ for its implementation. Let $W_i = \{w_k\}_{k=1}^{i-1}$ denote the set of intermediate results from all preceding steps. When a predefined function is invoked:
\begin{equation}
\label{eq:ci_pred}
c_i = f_i(\mathcal{D}_{cg}, \tilde{W}_i), \quad f_i \in \mathcal{F}_{pred},
\end{equation}
where $\tilde{W}_i \subseteq W_i$ contains selected results serving as input to the current step.

However, $\mathcal{F}_{pred}$ is inherently limited in scope, unable to handle ``out-of-predefined'' operations required by complex questions. This limitation motivates a flexible approach that can dynamically generate custom functions tailored to specific operational requirements.

\begin{table}[t]
\centering
\footnotesize
\renewcommand{\arraystretch}{1.05}
\begin{tabular}{@{}l l@{}}
\toprule
\textbf{Operation} & \textbf{Definition} \\
\midrule
\multicolumn{2}{@{}l}{\textit{Set Operations}} \\
Union       & $f_{\cup}(S_1, \ldots, S_n) \mapsto \bigcup_{i=1}^{n} S_i$ \\
Intersection& $f_{\cap}(S_1, \ldots, S_n) \mapsto \bigcap_{i=1}^{n} S_i$ \\
Difference  & $f_{-}(S_1, S_2) \mapsto S_1 \setminus S_2$ \\
\midrule
\multicolumn{2}{@{}l}{\textit{Algebraic Operations}} \\
Min / Max   & $f_{\min}(S) \mapsto \min(S)$; \ $f_{\max}(S) \mapsto \max(S)$ \\
Mean        & $f_{avg}(S) \mapsto \frac{1}{|S|}\sum_{x \in S} x$ \\
Count / Sum & $f_{cnt}(S) \mapsto |S|$; \ $f_{sum}(S) \mapsto \sum_{x \in S} x$ \\
\bottomrule
\end{tabular}

\caption{Predefined Calculation Operations}
\label{tab:support-ops}
\vspace{-1.2\baselineskip}

\end{table}

\subsection{CRAFT Module}

CRAFT extends CodeSTEP by dynamically generating custom functions for operations beyond $\mathcal{F}_{pred}$.

\paragraph{CRAFT Interface Function.}
To enable seamless invocation of CRAFT within CodeSTEP, we define an interface function $f_{craft}$ that serves as the bridge between the two modules:
\begin{equation}
\label{eq:fcraft}
f_{craft}: (\mathcal{T}_i, W_i, F_{exp,i}) \rightarrow w_i,
\end{equation}
where $\mathcal{T}_i$ is the task description for the current step, $W_i$ contains intermediate results from preceding steps, $F_{exp,i}$ is the expected function signature, and $w_i$ is the computed result. This interface function encapsulates the entire CRAFT execution process, allowing CodeSTEP to invoke the CRAFT module to perform complex custom operations through a unified calling convention.

With this interface, we define the extended function set available to CodeSTEP:
\begin{equation}
\label{eq:fext}
\mathcal{F} = \mathcal{F}_{pred} \cup \{f_{craft}\}.
\end{equation}

Each code step $c_i$ in the generated sequence $\mathcal{C}$ is now expressed as:
\begin{equation}
\label{eq:ci_complete}
c_i =
\begin{cases}
f_i(\mathcal{D}_{cg}, \tilde{W}_i), & f_i \in \mathcal{F}_{pred}, \\[0.5ex]
f_{craft}(\mathcal{T}_i, W_i, F_{exp,i}), & \text{otherwise}.
\end{cases}
\end{equation}

This formulation enables CodeSTEP to maintain a consistent format for stepwise code reasoning, where each $c_i$ uniformly represents a function call within the extended set $\mathcal{F}$.

\paragraph{Custom Function Generation.}
When $f_{craft}$ is invoked during execution, it triggers the CRAFT module to generate and execute a custom function. Internally, CRAFT utilizes an LLM $\mathcal{M}_{\theta'}$ with specialized prompt $\mathcal{P}_c$ to synthesize a self-contained custom function $\hat{f}_i$:
\begin{equation}
\label{eq:craft}
\mathcal{M}_{\theta'}(\mathcal{Q}, \mathcal{C}, \mathcal{T}_i, W_i, F_{exp,i}, \mathcal{P}_c) \rightarrow \hat{f}_i.
\end{equation}

The output of $f_{craft}$ is obtained by executing the generated function $\hat{f}_i$ from Equation \ref{eq:craft}:
\begin{equation}
\label{eq:fcraft_exec}
f_{craft}(\mathcal{T}_i, W_i, F_{exp,i}) := \textsc{Exec}(\hat{f}_i).
\end{equation}

Note that $\mathcal{Q}$, $\mathcal{C}$, and $\mathcal{P}_c$ are accessible as global context during execution, while $\mathcal{T}_i$, $W_i$, and $F_{exp,i}$ are step-specific parameters passed through the interface.

The input to CRAFT comprises five carefully designed components that convey $\mathcal{M}_\theta$'s understanding of the current step to $\mathcal{M}_{\theta'}$, enabling context-aware code generation: 
\textbf{(1) the original question $\mathcal{Q}$}, which provides the ultimate answering objective, enabling $\mathcal{M}_{\theta'}$ to consider whether the current step contributes to the final answer; 
\textbf{(2) the complete code sequence $\mathcal{C}$} generated by $\mathcal{M}_\theta$, representing the overall reasoning chain of thought for the task, which helps $\mathcal{M}_{\theta'}$ understand the role of the current step within the holistic reasoning process; 
\textbf{(3) the current task description $\mathcal{T}_i$}, articulated by $\mathcal{M}_\theta$ based on its understanding of the current step's requirements, thereby conveying the specific functional needs to $\mathcal{M}_{\theta'}$; 
\textbf{(4) the previous results $W_i$}, providing data formats and intermediate outputs from preceding steps that interact with $\mathcal{D}_{cg}$, enabling $\mathcal{M}_{\theta'}$ to generate functions with accurate input/output handling; 
\textbf{(5) the expected function signature $F_{exp,i}$}, a reference function name proposed by $\mathcal{M}_\theta$ that encapsulates its understanding of the required operation, helping bridge the comprehension gap between the two models regarding the current scenario.

\paragraph{Seamless Integration with CodeSTEP.}
The integration between CRAFT and CodeSTEP follows a well-defined workflow:

\noindent\textbf{(1) Code Generation with Delegation.} During code generation (Equation \ref{eq:code_gen}), when $\mathcal{M}_\theta$ encounters operations outside $\mathcal{F}_{pred}$, it generates $f_{craft}$ calls with appropriate parameters $(\mathcal{T}_i, W_i, F_{exp,i})$, as formulated in Equation \ref{eq:ci_complete}.

\noindent\textbf{(2) Deferred Execution.} The $f_{craft}$ calls remain as placeholders in $\mathcal{C}$ until execution time. This deferred execution strategy ensures that CRAFT has access to actual intermediate results $W_i$ computed from preceding steps.

\noindent\textbf{(3) Dynamic Function Synthesis.} Upon reaching an $f_{craft}$ call during execution, CRAFT synthesizes the custom function $\hat{f}_i$ via Equation \ref{eq:craft}. By leveraging $\mathcal{Q}$, $\mathcal{C}$, $\mathcal{T}_i$, $W_i$, and $F_{exp,i}$, CRAFT fully comprehends the context of the current step, enabling it to generate code that optimally addresses the step-specific problem.

\noindent\textbf{(4) Execution and Continuation.} The generated function $\hat{f}_i$ is executed, and its result $w_i$ is returned through $f_{craft}$ to the main CodeSTEP execution flow. Subsequent steps can then utilize $w_i$ as part of their input $W_{i+1}$, maintaining the sequential reasoning chain.

\subsection{Code Verification and Execution}

\paragraph{Executability Verification.}
To ensure the generated code is syntactically correct, we verify each code block $\mathcal{B}$ (either $\mathcal{C}$ from CodeSTEP or $\hat{f}$ from CRAFT) through a Python interpreter:
\begin{equation}
\label{eq:verify}
\textsc{Verify}(\mathcal{B}) =
\begin{cases}
1, & \text{if } \mathcal{B} \text{ is valid}, \\
0, & \text{otherwise}.
\end{cases}
\end{equation}

When verification fails, we regenerate $\mathcal{B}$ with identical input up to $T$ attempts. The verified code $\mathcal{B}^*$ is obtained at the first successful attempt, or defaults to $\mathcal{B}^{(T)}$ if all attempts fail.

\paragraph{Complete Execution Process.}
The verified code $\mathcal{C}^* = \{c_i\}_{i=1}^{n}$ is executed as a complete unit in a single run, where each code step $c_i$ is processed sequentially within the execution. Whether invoking a predefined function or $f_{craft}$, each step produces an intermediate result:
\begin{equation}
\label{eq:step_exec}
w_i = \textsc{Exec}(c_i, \mathcal{D}_{cg}, W_i).
\end{equation}

For steps invoking $f_{craft}$, the execution internally triggers CRAFT to generate $\hat{f}_i$ (Equation \ref{eq:craft}), verifies its executability (Equation \ref{eq:verify}), and returns the result through the interface (Equation \ref{eq:fcraft_exec}).

These intermediate results form a sequential reasoning chain, where each $w_i$ is accumulated into $W_{i+1} = W_i \cup \{w_i\}$ for subsequent steps. The final result $w_n$ directly corresponds to the answer $\mathcal{A}$:
\begin{equation}
\label{eq:final_exec}
\textsc{Exec}(\mathcal{C}^*, \mathcal{D}_{cg}) \rightarrow w_n = \mathcal{A}.
\end{equation}


\section{Experiments}

We conduct comprehensive experiments to answer the following four key research questions:
\textbf{RQ1}: How effective is CRAFTQA in complex structured data reasoning tasks, particularly in ``out-of-predefined'' scenarios?
\textbf{RQ2}: While achieving improvements in complex reasoning, does CRAFTQA maintain competitive performance on standard reasoning tasks across different types of structured data?
\textbf{RQ3}: How generalizable is CRAFTQA when applied to diverse backbone LLMs with varying capabilities and scales?
\textbf{RQ4}: How does each component in the CRAFTQA framework contribute to the overall performance?

\subsection{Experimental Setup}

\paragraph{Datasets and Evaluation Metrics.}
For \textbf{RQ1}, we use two challenging sub-datasets from TableBench \cite{wu2025tablebench}: Fact Checking (FC) and Numerical Reasoning (NR), along with WikiSQL-E, a dataset we constructed by extracting question-answer pairs from WikiSQL \cite{zhong2017seq2sql} that may involve ``out-of-predefined'' operations (details in Appendix \ref{sec:wikisql_E_construction}).
We use Denotation Accuracy (DA) \cite{jiang2023structgpt} as the primary metric, and compute an Overall metric as the weighted average of DA: 
\begin{equation}
\text{Overall} = \frac{\sum_{i=1}^{K} n_i \cdot DA_i}{\sum_{i=1}^{K} n_i},
\end{equation}
where $K$ is the number of datasets, $n_i$ and $DA_i$ denote the sample size and DA of the $i$-th dataset.

To assess the correctness of answers requiring ``out-of-predefined'' functions, we propose Calling Denotation Accuracy (\textbf{CDA}):
\begin{equation}
\text{CDA} = \frac{N_{calling\_correct}}{N_{calling}},
\end{equation}
where $N_{calling}$ is the number of questions calling ``out-of-predefined'' functions, and $N_{calling\_correct}$ denotes the number that are answered correctly. Detailed definitions are in Appendix \ref{sec:CDA_definition}.

For \textbf{RQ2}, we use WebQSP \cite{yih2016value} (KGQA), WikiSQL (TableQA), and CronQuestions \cite{saxena2021question} (TKGQA), with Hit@1 for WebQSP and CronQuestions, DA for WikiSQL.

\begin{figure}[t]
    \centering
    \includegraphics[width=0.95\linewidth, trim=0 0 0 0, clip]{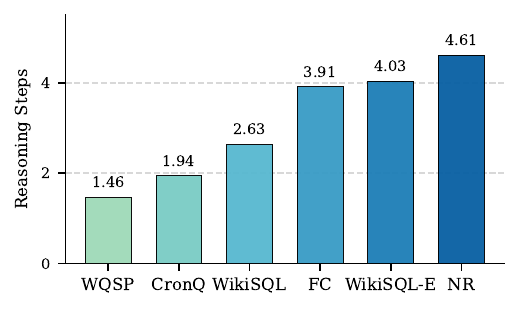}
    
    \vspace{-0.35cm}
    
    \caption{\textbf{Average Reasoning Steps} across Datasets.}
    \label{fig:reasoning_steps}
    
    \vspace{-0.5cm}
\end{figure}

To quantify the reasoning complexity of datasets, inspired by TableBench \cite{wu2025tablebench}, we adopt Average Reasoning Steps as a metric. As shown in Figure \ref{fig:reasoning_steps}, \textbf{RQ1} datasets (WikiSQL-E, TableBench-FC, and TableBench-NR) exhibit higher average reasoning steps, indicating greater complexity. In contrast, \textbf{RQ2} datasets (WebQSP, CronQuestions, and WikiSQL) have lower average reasoning steps, representing standard reasoning tasks. Detailed statistics are provided in Appendix \ref{sec:reasoning_complexity_stat}.

For \textbf{RQ3}, we evaluate on TableBench (FC and NR) using CDA, DA, and F1. More dataset statistics are in Appendix \ref{sec:appendix_dataset_metric}.

\paragraph{Baselines.}
For RQ1 and RQ2, we compare CRAFTQA against advanced unified structured data reasoning methods: PoT \cite{chen2022program}, which generates executable programs for reasoning; StructGPT \cite{jiang2023structgpt}, which employs iterative reading-then-reasoning; Readi \cite{cheng2024call}, which edits reasoning paths via environmental feedback; and TrustUQA \cite{zhang2025trustuqa}, which uses unified Condition Graph for querying. For the ``Out-of-Predefined'' scenarios experiments (RQ1), we adopt TrustUQA as the primary baseline, as it represents the \textbf{state-of-the-art} among published methods.

\paragraph{Implementation Details.}
Our experiments employ various LLMs as inference engines, including GPT \cite{achiam2023gpt}, LLaMA \cite{dubey2024llama}, DeepSeek \cite{liu2024deepseek}, Gemini \cite{team2023gemini}, and Qwen \cite{bai2023qwen} series. We adopt a self-consistency strategy with 5 samples, set the maximum retry attempts $T{=}3$, and use Sentence-BERT \cite{reimers2019sentence} to semantically align entities. Further details are provided in Appendix~\ref{sec:appendix_implement_detail}.

\subsection{CRAFTQA for Complex Reasoning (RQ1)}

\begin{table}[t]
    \centering
    \small 
    \renewcommand{\arraystretch}{1.1} 
    \setlength{\tabcolsep}{2pt}

    \resizebox{\columnwidth}{!}{%
    \begin{tabular}{l c c c c}
        \toprule
        \multirow{2}{*}{\textbf{Methods}} & \multicolumn{2}{c}{\textbf{TableBench}} & \textbf{WikiSQL-E} & \textbf{Overall} \\
        \cmidrule(lr){2-3} \cmidrule(lr){4-4} \cmidrule(lr){5-5}
         & \footnotesize FC & \footnotesize NR & \footnotesize DA(\%) & \footnotesize Avg. \\
        \midrule
        
        \multicolumn{5}{l}{\textbf{PoT}~\cite{chen2022program}} \\
        \hspace{1em}-- GPT-3.5-Turbo & 54.2 & 36.4 & 25.0 & 29.4 \\
        \hspace{1em}-- GPT-4o-mini   & 41.7 & 36.9 & 25.0 & 28.8 \\
        \hspace{1em}-- GPT-4o        & 51.0 & 43.7 & 27.4 & 32.6 \\
        \addlinespace[0.2em] 
        
        \multicolumn{5}{l}{\textbf{StructGPT}~\cite{jiang2023structgpt}} \\
        \hspace{1em}-- GPT-3.5-Turbo & 56.3 & 23.0 & 44.0 & 39.7 \\
        \hspace{1em}-- GPT-4o-mini   & 58.3 & 24.2 & 38.8 & 36.5 \\
        \hspace{1em}-- GPT-4o        & 63.5 & 42.2 & 43.5 & 44.3 \\
        \addlinespace[0.2em]
        
        \multicolumn{5}{l}{\textbf{Readi}~\cite{cheng2024call}} \\
        \hspace{1em}-- GPT-3.5-Turbo & 51.0 & 33.6 & 45.1 & 42.7 \\
        \hspace{1em}-- GPT-4o-mini   & 55.2 & 38.4 & 37.5 & 38.7 \\
        \hspace{1em}-- GPT-4o        & 62.5 & 49.5 & 51.3 & 51.5 \\
        \addlinespace[0.2em]
        
        \multicolumn{5}{l}{\textbf{TrustUQA}~\cite{zhang2025trustuqa}} \\
        \hspace{1em}-- GPT-3.5-Turbo & 50.0 & 20.2 & \textbf{70.0} & 57.1 \\
        \hspace{1em}-- GPT-4o-mini   & 55.2 & 21.7 & 74.5 & 61.0 \\
        \hspace{1em}-- GPT-4o        & 62.5 & 29.6 & 80.1 & 67.2 \\
        \addlinespace[0.2em]
        
        \midrule
        \rowcolor{mygray} \multicolumn{5}{l}{\textbf{CRAFTQA(Ours)}} \\
        
        \rowcolor{mygray} \multicolumn{5}{l}{\textit{\hspace{0.5em}Open-source Models}} \\
        \rowcolor{mygray} \hspace{1.5em}-- Qwen2.5-7B   & 57.9 & 35.4 & 65.9 & 58.3 \\
        \rowcolor{mygray} \hspace{1.5em}-- LLaMA-3.1-8B & 57.3 & 31.1 & 76.0 & 64.4 \\
        
        \rowcolor{mygray} \multicolumn{5}{l}{\textit{\hspace{0.5em}Closed-source Models}} \\
        \rowcolor{mygray} \hspace{1.5em}-- GPT-3.5-Turbo & \textbf{61.5} & \textbf{38.6} & \underline{67.8} & \textbf{60.6} \\
        \rowcolor{mygray} \hspace{1.5em}-- GPT-4o-mini   & \textbf{64.6} & \textbf{40.7} & \textbf{79.0} & \textbf{69.2} \\
        \rowcolor{mygray} \hspace{1.5em}-- GPT-4o        & \textbf{68.8} & \textbf{51.3} & \textbf{85.6} & \textbf{76.6} \\
        
        \bottomrule
    \end{tabular}%


    }

    \caption{\textbf{Complex reasoning} performance comparison.
    }
    \label{tab:complex_reasoning_performance}
    \vspace{-1.2\baselineskip}

\end{table}

To answer \textbf{RQ1}, we evaluate CRAFTQA on three complex reasoning datasets: TableBench-FC, TableBench-NR, and WikiSQL-E, which demand significantly more reasoning steps (Figure \ref{fig:reasoning_steps}).

Table \ref{tab:complex_reasoning_performance} compares CRAFTQA with advanced unified structured data reasoning methods, including PoT \cite{chen2022program}, StructGPT \cite{jiang2023structgpt}, Readi \cite{cheng2024call}, and TrustUQA \cite{zhang2025trustuqa}. Under the same backbone LLM, CRAFTQA consistently achieves the best performance across nearly all datasets. With GPT-4o, CRAFTQA attains 68.8\%, 51.3\%, and 85.6\% DA on FC, NR, and WikiSQL-E respectively, significantly outperforming all baselines. Similar advantages are observed with GPT-4o-mini. The only exception occurs with GPT-3.5-Turbo on WikiSQL-E, where TrustUQA achieves 70.0\% compared to our 67.8\%. This is expected since code-based methods inherently require sufficient backbone LLM capabilities \cite{chen2022program}, and GPT-3.5-Turbo's limited reasoning ability affects the performance. Nevertheless, CRAFTQA still achieves the best Overall metric (60.6\%) with GPT-3.5-Turbo.

We further explore CRAFTQA with smaller open-source models. Notably, CRAFTQA-LLaMA-3.1-8B (64.4\%) and CRAFTQA-Qwen2.5-7B (58.3\%) even outperform baselines using larger closed-source models, such as Readi-GPT-4o (51.5\%), StructGPT-GPT-4o (44.3\%), and PoT-GPT-4o (32.6\%), demonstrating CRAFTQA's superiority in complex reasoning scenarios.

\subsection{CRAFTQA for ``Out-of-Predefined'' Scenarios (RQ1)}

\begin{figure}[t!]
    \centering
    \vspace{-0.5ex}
    \includegraphics[width=0.95\linewidth, height=0.28\textheight, keepaspectratio]{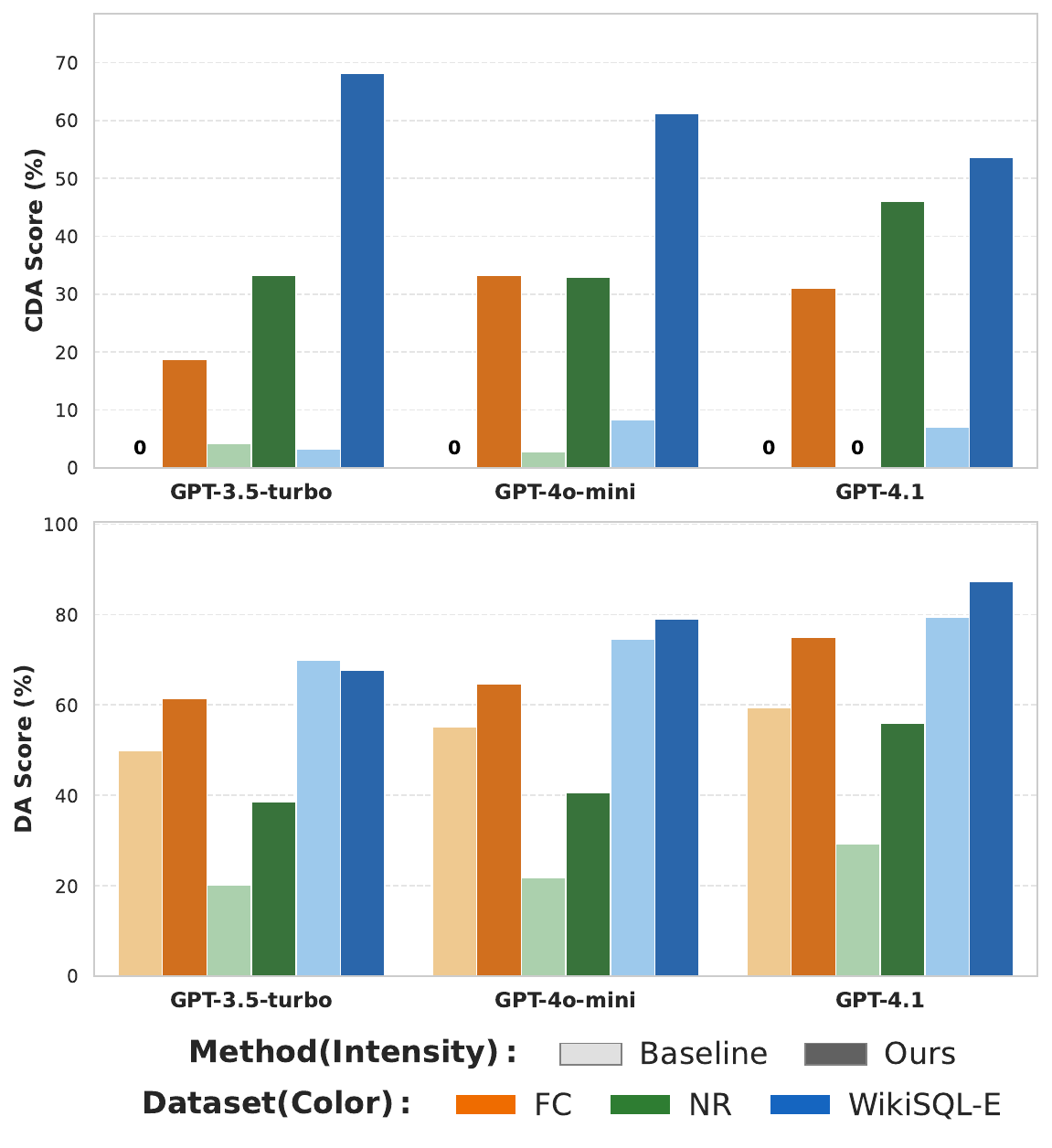}
    \vspace{-2ex}
    \caption{\textbf{Calling Denotation Accuracy (CDA) and Denotation Accuracy (DA) comparison between CRAFTQA and the state-of-the-art baseline.} Detailed results are in Table~\ref{tab:CRAFTQA_with_TrustUQA_CDA_results}.}
    \label{fig:CDA_metrics_comparison_with_sotaTrustUQA}
    \vspace{-2.5ex}
\end{figure}

To investigate CRAFTQA's effectiveness in handling ``out-of-predefined'' scenarios, we compare it against TrustUQA~\cite{zhang2025trustuqa}, the current published state-of-the-art method on unified structured data QA. We evaluate on WikiSQL-E and TableBench (FC and NR) using GPT-3.5-Turbo, GPT-4o-mini, and GPT-4.1 as backbone LLMs, with results shown in Figure~\ref{fig:CDA_metrics_comparison_with_sotaTrustUQA}.

To quantify performance on ``out-of-predefined'' scenarios, we track questions that cannot be solved by predefined functions and calculate the Calling Denotation Accuracy (CDA) for these cases. As shown in Figure~\ref{fig:CDA_metrics_comparison_with_sotaTrustUQA}, \textbf{CRAFTQA consistently achieves substantially higher CDA across all datasets and backbone LLMs}. For instance, on WikiSQL-E with GPT-4.1, CRAFTQA achieves a CDA of 53.57\% compared to TrustUQA's 6.98\%, demonstrating CRAFT's advantage in handling tasks requiring ``out-of-predefined'' functions.

Furthermore, improvements in ``out-of-predefined'' scenarios contribute to substantial overall performance gains. CRAFTQA \textbf{significantly outperforms} TrustUQA in both DA and F1 metrics across nearly all settings. For example, on the Numerical Reasoning dataset with GPT-4.1, CRAFTQA achieves a DA of 56.06\% versus TrustUQA's 29.29\%, an improvement of 26.77 percentage points. These results demonstrate our framework's superiority in complex reasoning tasks, where effective handling of ``out-of-predefined'' scenarios plays a crucial role.

\subsection{CRAFTQA for Standard Reasoning(RQ2)}


Having demonstrated CRAFTQA's superiority in complex ``out-of-predefined'' scenarios (RQ1), we now investigate whether CRAFTQA maintains competitive performance on standard reasoning tasks across heterogeneous structured data sources. We compare CRAFTQA against state-of-the-art unified structured data QA methods on three representative datasets: WebQSP (Knowledge Graph), WikiSQL (Table), and CronQuestions (Temporal Knowledge Graph), as shown in Table~\ref{tab:heterogeneous_performance}.

As shown in Table~\ref{tab:heterogeneous_performance}, CRAFTQA achieves the best performance on both WebQSP (85.20\% Hit@1) and WikiSQL (86.10\% DA), outperforming TrustUQA~\cite{zhang2025trustuqa} by 1.70 and 0.40 percentage points, respectively. On CronQuestions, CRAFTQA achieves 97.10\% accuracy, comparable to TrustUQA's 97.20\%. This marginal difference can be attributed to the inherent simplicity of TKG-based tasks: as illustrated in Figure~\ref{fig:reasoning_steps}, CronQuestions requires fewer reasoning steps and TKG data involves limited operation variety, resulting in rare ``out-of-predefined'' scenarios that leave limited room for CRAFT to demonstrate its advantage. Nevertheless, CRAFTQA still achieves near-optimal performance, demonstrating its robustness across varying task complexities.

These results demonstrate that \textbf{CRAFTQA maintains comparable or even superior standard reasoning capabilities while significantly enhancing complex reasoning performance}, and is applicable to diverse structured data sources including KG, Table, and TKG.

\begin{table}[t]
    \centering
    \renewcommand{\arraystretch}{1.27} 
    \setlength{\tabcolsep}{1pt}

    \resizebox{\columnwidth}{!}{%
    \begin{tabular}{l c c c}
        \toprule
        \multirow{2}{*}{\textbf{Methods}} & \textbf{WebQSP} & \textbf{WikiSQL} & \textbf{CronQ} \\
         & \text{Hit@1} & \text{DA(\%)} & \text{Hit@1} \\
        \midrule
        
        PoT~\cite{chen2022program} & 14.70 & 44.42 & 57.63 \\

        StructGPT~\cite{jiang2023structgpt} & 69.60 & 57.90 & -- \\
        Readi~\cite{cheng2024call}          & 74.30 & 64.70 & -- \\
        TrustUQA~\cite{zhang2025trustuqa}   & 83.50 & 85.70 & \textbf{97.20} \\
        \midrule 
        
        \rowcolor{mygray} 
        \textbf{CRAFTQA (Ours)} & \textbf{85.20} & \textbf{86.10} & \underline{97.10} \\
        
        \bottomrule
    \end{tabular}%

    }
        
    \caption{\textbf{Standard reasoning performance across heterogeneous structured data sources.} }
    \label{tab:heterogeneous_performance}

    \vspace{-1.2\baselineskip}

\end{table}

\subsection{Generalization of CRAFTQA (RQ3)}

To evaluate CRAFTQA's generalization, we assess its performance across different backbone LLMs. We experiment on TableBench's Fact Checking and Numerical Reasoning datasets using 4 open-source models (LLaMA-3.1-8B, Qwen2.5-7B, Qwen3-32B, DeepSeek-V3) and 8 closed-source models (Qwen-Max, Gemini-2.5-Flash/Pro, GPT-3.5, GPT-4o-mini, GPT-4o, GPT-5-mini, o4-mini). Detailed results are provided in Appendix Table~\ref{tab:craftqa_generalization}.

Figure~\ref{fig:CRAFT_generalization} illustrates the relationship between CDA (line) and overall metrics DA/F1 (bars) across backbone LLMs, revealing two key observations:

First, CRAFTQA's performance scales with the backbone LLM's capability. On Fact Checking, DA improves from 61.5\% (GPT-3.5) to 68.8\% (GPT-4o), and further to 83.3\% (o4-mini), suggesting that \textbf{CRAFTQA is well-positioned to benefit from future LLM advancements}.

Second, within the same LLM family, CDA trends exhibit strong consistency with DA and F1, validating that overall performance gains directly stem from CRAFT's capability in handling ``out-of-predefined'' scenarios.

Notably, as shown in Figure~\ref{fig:CRAFT_generalization}, the dashed line represents TrustUQA's DA with GPT-4o. Combined with Table~\ref{tab:complex_reasoning_performance}, we observe that CRAFTQA enables small-parameter open-source models to achieve competitive or superior performance compared to advanced baselines using large closed-source models (GPT-4o). These results confirm CRAFTQA's strong generalization and scalability across diverse backbone LLMs.

\begin{figure}[t]
    \centering
    \includegraphics[width=1.0\columnwidth]{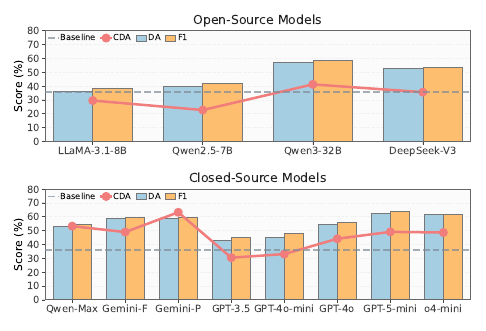}
    \caption{\textbf{Generalization of CRAFTQA} across backbone LLMs. 
    }
    \label{fig:CRAFT_generalization}
\end{figure}

\subsection{Ablation Study (RQ4)}

To validate the contribution of each module in CRAFTQA, we conduct ablation experiments on TableBench's Fact Checking and Numerical Reasoning datasets using GPT-4o, and on WikiSQL-E using GPT-4.1, with results presented in Table~\ref{tab:ablation_combined}.

\begin{table}[t]
  \centering
  \small
  \setlength{\tabcolsep}{4pt}
  \renewcommand{\arraystretch}{1.15}
  
  \resizebox{\linewidth}{!}{
    \begin{tabular}{@{}l cccccc@{}}
      \toprule
      \multirow{2}{*}{\textbf{Methods}}
        & \multicolumn{2}{c}{\textbf{FC}} 
        & \multicolumn{2}{c}{\textbf{NR}}
        & \multicolumn{2}{c}{\textbf{WikiSQL-E}} \\
      \cmidrule(lr){2-3} \cmidrule(lr){4-5} \cmidrule(l){6-7}
        & DA & F1 & DA & F1 & DA & F1 \\
      \midrule
      \textbf{CRAFTQA}
        & \textbf{68.8} & \textbf{71.6}
        & \textbf{51.3} & \textbf{51.9}
        & \textbf{87.3} & \textbf{87.6} \\
      w/o CRAFT
        & 65.3 & 67.7
        & 45.9 & 46.3
        & 84.5 & 84.6 \\
      w/o CRAFT\&CodeSTEP
        & 59.4 & 63.2
        & 18.4 & 19.7
        & 79.8 & 80.0 \\
      \bottomrule
    \end{tabular}
  }
  \caption{\textbf{Ablation Study results} across Fact Checking (FC), Numerical Reasoning (NR), and WikiSQL-E.}
  \label{tab:ablation_combined}
    \vspace{-1.2\baselineskip}
\end{table}
 
Removing either the CRAFT or CodeSTEP module significantly degrades performance. For instance, DA on Numerical Reasoning drops from 51.3\% (CRAFTQA) to 45.9\% (w/o CRAFT) and further to 18.4\% (w/o CRAFT\&CodeSTEP). These results validate that both modules are crucial: CodeSTEP provides core reasoning structure while CRAFT offers flexibility for handling ``out-of-predefined'' scenarios, and their synergy further enhances performance on complex reasoning tasks.

\section{Conclusion}

In this paper, we present CRAFTQA, an efficient and flexible code-driven framework for unified structured data question answering, comprising two core modules: CodeSTEP and CRAFT. 
CodeSTEP generates and executes code sequences to answer natural language questions directly, and works seamlessly with the CRAFT module, which can effectively handles ``out-of-predefined'' scenarios. 
Experiments on diverse datasets demonstrate our framework's effectiveness, particularly on complex reasoning tasks, and validate its generalization capabilities. 
As LLMs continue to evolve, CRAFTQA is well-positioned to benefit from future LLMs advancements. Looking forward, we plan to extend our framework to more data forms, and further enhance its capabilities.

\section*{Limitations}

While CRAFTQA demonstrates significant improvements in complex reasoning scenarios, we acknowledge several limitations. First, the performance gains on simpler standard reasoning tasks are modest, these tasks typically involve fewer ``out-of-predefined'' operations, which limits the opportunities for the CRAFT module to fully leverage its dynamic reasoning capabilities. Second, as a code-based framework, CRAFTQA may present challenges for LLMs with limited code generation abilities, potentially restricting its applicability to models with weaker programming proficiency. These limitations present opportunities for future research to extend the framework's applicability across broader reasoning scenarios.

\section*{Acknowledgments}

This work is founded by National Natural Science Foundation of China (NSFC62306276/NSFCU23B2055), New Generation Artificial Intelligence-National Science and Technology Major Project 2030 (2025ZD0122800), Yongjiang Talent Introduction Programme (2022A-238-G), and Fundamental Research Funds for the Central Universities (226-2023-00138). This work was supported by Ant Group. 


\bibliography{custom}

\appendix

\section{Datasets and Evaluation Metrics}
\label{sec:appendix_dataset_metric}


We evaluate on multiple standard QA datasets spanning different structured data types,
\textbf{WebQSP} \cite{yih2016value} is a KGQA dataset requiring reasoning over Freebase. We use Hit@1 as the evaluation metric.
\textbf{CronQuestions} \cite{saxena2021question} is a temporal knowledge graph question answering (TKGQA) dataset designed to evaluate models’ ability to reason over time-dependent facts. Each question is associated with entities, relations, and timestamps, requiring temporal reasoning across evolving knowledge graphs. We use Hit@1 as the evaluation metric.
\textbf{WikiSQL} \cite{zhong2017seq2sql} is a table QA dataset requiring to answer questions over Wikipedia tables. We use Denotation Accuracy (DA)\cite{jiang2023structgpt} as the evaluation metric.
\textbf{TableBench} \cite{wu2025tablebench} is a comprehensive and complex benchmark for table reasoning. We focus on two challenging subtasks, Fact Checking (FC) and Numerical Reasoning (NR), using Denotation Accuracy (DA) as the evaluation metric. 
Statistics of datasets are shown in Table \ref{tab:data_stats}.

\begin{table}[htbp]
    \centering
    \resizebox{\linewidth}{!}{
    \begin{tabular}{lrcr}
        \toprule
        \textbf{Dataset} & \textbf{\# QA Pairs} & \textbf{Struct. Data Vol.} & \textbf{Type} \\
        \midrule
        WebQSP & 1,639 & Retrieved Ver. & KG \\
        CronQuestions & 30,000 & 5,000 triples & TKG \\
        WikiSQL & 15,878 & 5,230 tables & Table \\
        TableBench-FC & 96 & 96 tables & Table \\
        TableBench-NR & 396 & 396 tables & Table \\
        WikiSQL-E & 1,190 & 925 tables & Table \\
        \bottomrule
    \end{tabular}%
    }
    
    \caption{\textbf{Statistics of datasets used in evaluation.} We report the number of Question-Answer (QA) pairs, the volume of structured data, and the target data type (KG: Knowledge Graph, TKG: Temporal KG).}
    \label{tab:data_stats}

    \vspace{-1.2\baselineskip}

\end{table}

\subsection{WikiSQL-E Construction}
\label{sec:wikisql_E_construction}

To better evaluate the performance of CRAFTQA in complex reasoning and ``out-of-predefined'' scenarios, we constructed a specialized dataset by extracting questions from WikiSQL that could not be well solved using only predefined functions, and used the extracted questions to construct a dataset named \textbf{WikiSQL-E}.

The construction of WikiSQL-E follows a systematic 4-step automated process, designed to identify questions that genuinely require operations beyond predefined function sets:

\paragraph{Step 1: Compile a Unified Predefined Function List.}
We systematically reviewed the predefined operations of existing state-of-the-art unified methods, including StructGPT~\cite{jiang2023structgpt}, Readi~\cite{cheng2024call}, and TrustUQA~\cite{zhang2025trustuqa}, and consolidated a comprehensive predefined function list that covers virtually all ``predefined'' executable operations shared across these methods.
The compiled function list includes:
(1) \texttt{``get\_information''}: retrieves information by querying a data source using specified relations and entities;
(2) set operations: \texttt{``set\_union''}, \texttt{``set\_intersection''}, and \texttt{``set\_difference''};
and (3) algebraic operations: \texttt{``Min''}, \texttt{``Max''}, \texttt{``Mean''}, \texttt{``Count''}, and \texttt{``Sum''}.

\paragraph{Step 2: Add a Placeholder Interface.}
We additionally introduced an \texttt{``out\_of\_predefined()''} placeholder function, which does not perform actual data processing at this stage, to simulate a callable interface for operations beyond the predefined function list. This enables the screening model to explicitly signal when predefined operations are insufficient for a given question.

\paragraph{Step 3: Automated LLM Screening.}
Using GPT-3.5-Turbo as the backbone LLM, we processed the entire WikiSQL dataset with prompts instructing the model to prioritize predefined functions and only invoke \texttt{``out\_of\_predefined()''} when no suitable predefined operation exists. This design allows the LLM to assess whether predefined operations are sufficient based on the question semantics, effectively identifying questions that require complex reasoning operations beyond the standard predefined function set.

\paragraph{Step 4: Filtering and Dataset Formation.}
QA pairs whose generated code invoked \texttt{``out\_of\_predefined()''} at least once were extracted to form the WikiSQL-E dataset. This filtering process yielded 1,190 QA pairs spanning 925 tables, as reported in Table~\ref{tab:data_stats}.

This automated construction process ensures that WikiSQL-E is not manually curated but systematically identified through an objective screening mechanism, capturing questions that are genuinely challenging for predefined-function-only methods.

We primarily use Denotation Accuracy (DA) and F1 scores to evaluate the overall performance of the methods. Additionally, we introduce a new metric called Calling Denotation Accuracy (\textbf{CDA}), which is designed to evaluate the correctness of the answers produced specifically by the ``out-of-predefined'' functions and we will define CDA in more detail below. These metrics help validate the effectiveness of the CRAFT module.

Experiments on the constructed dataset WikiSQL-E effectively explore our framework's improvements and capabilities in complex reasoning scenarios and provide a valuable dataset and baseline for future research in this area.

\subsection{Definition of Calling Denotation Accuracy}
\label{sec:CDA_definition}

The Calling Denotation Accuracy (\textbf{CDA}) metric is designed to evaluate the correctness of answers generated for questions that have called ``out-of-predefined'' functions. We utilize the self-consistency \cite{wang2022self} strategy in experiments. Specifically, for each question, the LLM generates $n$ distinct reasoning paths, and the final answer is determined by a majority vote among them. In our experiments, we set $n=5$.

Let $Q_{total}$ be the set of all questions in a given dataset. A question $q$ is considered to have called ``out-of-predefined'' function if at least one of its $n=5$ generated reasoning paths calls an ``out-of-predefined'' function. Let $Q_{calling}$ be the subset of such questions from $Q_{total}$, and $N_{calling}$ be its size.

For each question $q_{calling}$ in the $Q_{calling}$ subset, we first identify all reasoning paths that have called ``out-of-predefined'' functions. Assume that for each question $q_{calling}$, there are $k$ such paths (where $1 \leq k \leq n$). We then perform a majority vote only among the $k$ answers generated by these reasoning paths to determine a single, definitive answer $a_{calling}$ for that question. Let $N_{calling\_correct}$ be the number of questions in $Q_{calling}$ for which $a_{calling}$ matches the ground-truth answer. The Calling Denotation Accuracy (\textbf{CDA}) is then defined as:
\begin{equation}
\text{CDA} = \frac{N_{calling\_correct}}{N_{calling}}.
\end{equation}

\begin{table}[htbp]
    \centering
    \resizebox{\linewidth}{!}{
    \begin{tabular}{llcc}
        \toprule
        \multirow{2}{*}{\textbf{Dataset}} & \multirow{2}{*}{\textbf{Scenario}} & \multicolumn{2}{c}{\textbf{Reasoning Steps}} \\
        \cmidrule(lr){3-4}
         &  & \textbf{Avg.} & \textbf{Max.} \\
        \midrule
        WebQSP & KGQA & 1.46 & 4 \\
        CronQuestions & Temporal KGQA & 1.94 & 6 \\
        WikiSQL & TableQA & 2.63 & 9 \\
        \midrule
        TableBench-FC & Complex TableQA & 3.91 & 9 \\
        TableBench-NR & Complex TableQA & 4.61 & 15 \\
        WikiSQL-E & Complex TableQA & 4.03 & 9 \\
        \bottomrule
    \end{tabular}%
    }
    \caption{\textbf{Reasoning complexity statistics.} We list the applicable task scenarios for each dataset, and statistically analyze the average and maximum reasoning steps required to answer the questions in each dataset.}
    \label{tab:reasoning_complexity}

    \vspace{-1.2\baselineskip}

\end{table}

\begin{table*}[t]
  \centering
  \small %
  \renewcommand{\arraystretch}{1.25} 
  \setlength{\tabcolsep}{12pt}
  
  \begin{tabular}{@{}ll ccc@{}}
    \toprule
    \multirow{2}{*}{\textbf{Model}} & \multirow{2}{*}{\textbf{Dataset}} & \textbf{CDA (\%)} & \textbf{DA (\%)} & \textbf{F1 (\%)} \\
    \cmidrule(lr){3-5}
    & & (TrustUQA / Ours) & (TrustUQA / Ours) & (TrustUQA / Ours) \\
    \midrule
    
    \multirow{4}{*}{GPT-3.5-Turbo} 
      & WikiSQL-E           & 3.32 / 68.24 $\uparrow$  & 70.00 / 67.79            & 70.40 / 68.28 \\
      & FactChecking        & 0.00 / 18.75 $\uparrow$  & 50.00 / 61.46 $\uparrow$ & 56.61 / 65.44 $\uparrow$ \\
      & Numerical Reasoning & 4.23 / 33.33 $\uparrow$  & 20.20 / 38.63 $\uparrow$ & 20.87 / 40.10 $\uparrow$ \\
    \midrule
    
    \multirow{4}{*}{GPT-4o-mini} 
      & WikiSQL-E           & 8.33 / 61.29 $\uparrow$  & 74.54 / 79.04 $\uparrow$ & 75.06 / 79.66 $\uparrow$ \\
      & FactChecking        & 0.00 / 33.33 $\uparrow$  & 55.21 / 64.58 $\uparrow$ & 58.39 / 70.19 $\uparrow$ \\
      & Numerical Reasoning & 2.82 / 33.00 $\uparrow$  & 21.72 / 40.66 $\uparrow$ & 22.73 / 42.57 $\uparrow$ \\
    \midrule
    \multirow{3}{*}{GPT-4.1} 
      & FactChecking        & 0.00 / 31.03 $\uparrow$  & 59.38 / 75.00 $\uparrow$ & 62.64 / 78.23 $\uparrow$ \\
      & Numerical Reasoning & 0.00 / 45.99 $\uparrow$  & 29.29 / 56.06 $\uparrow$ & 28.89 / 56.37 $\uparrow$ \\
      & WikiSQL-E           & 6.98 / 53.57 $\uparrow$ & 79.44 / 87.34 $\uparrow$ & 79.52 / 87.58 $\uparrow$ \\
      
    \bottomrule
  \end{tabular}
  
  \caption{Performance comparison on different datasets across GPT-3.5-Turbo, GPT-4o-mini, and GPT-4.1. The results are presented in the format of \textit{TrustUQA / Ours}. The $\uparrow$ symbol indicates that our method outperforms the baseline.}
  \label{tab:CRAFTQA_with_TrustUQA_CDA_results}
    \vspace{-1.2\baselineskip}

\end{table*}

\subsection{Quantification of Dataset Reasoning Complexity}
\label{sec:reasoning_complexity_stat}

To quantify the reasoning complexity of datasets, inspired by TableBench \cite{wu2025tablebench}, we adopt \textbf{Reasoning Steps} as the metric for measuring the reasoning difficulty of each dataset. The reasoning steps represent the number of intermediate operations required to derive the final answer from a given question.

For a dataset $\mathcal{D}$ containing $M$ question-answer pairs, let $step_i$ denote the reasoning steps required for the $i$-th question. We define the \textbf{Average Reasoning Steps} and \textbf{Maximum Reasoning Steps} as follows:
\begin{equation}
\text{Avg. Reasoning Steps} = \frac{1}{M} \sum_{i=1}^{M} step_i,
\end{equation}
\begin{equation}
\text{Max. Reasoning Steps} = \max_{i \in {1, \ldots, M}} step_i.
\end{equation}

Table \ref{tab:reasoning_complexity} presents the reasoning complexity statistics for all datasets used in our experiments. WebQSP, CronQuestions, and WikiSQL are relatively simple datasets with lower average reasoning steps (ranging from 1.46 to 2.63), which are employed to investigate the performance of our method on standard reasoning scenarios across different types of structured data. In contrast, TableBench-FC, TableBench-NR, and WikiSQL-E are more complex datasets with higher average reasoning steps (ranging from 3.91 to 4.61), which are utilized to evaluate the effectiveness of our method in complex reasoning scenarios.

\section{Other Implementation Details}
\label{sec:appendix_implement_detail}

All experiments are conducted on a server equipped with an Intel(R) Xeon(R) Gold 6148 CPU and three NVIDIA A100-SXM4-40GB GPUs, running on the Ubuntu 20.04.6 LTS operating system.

We use GPT-3.5 (GPT-3.5-Turbo) as the backbone LLM for the unified method on the WebQSP dataset, use GPT-4o-mini as the backbone LLM for the unified method on the CronQuestions and WikiSQL dataset. We use three different backbone LLMs, GPT-3.5, GPT-4o-mini, and GPT-4o, for experiments on different unified methods on the TableBench and WikiSQL-E dataset. 
And we use 4 open-source models (LLaMA-3.1-8B-Instruct, Qwen2.5-7B-Instruct, Qwen3-32B, DeepSeek-V3) and 8 closed-source models (Qwen-Max, Gemini-2.5-Flash, Gemini-2.5-Pro, GPT-3.5-Turbo, GPT-4o-mini, GPT-4o, GPT-5-mini, o4-mini) in the experiment that evaluate the generalization capabilities of CRAFTQA. 
For all LLM-based experiments, we use the self-consistency strategy of 5 times and use SentenceBERT~\cite{reimers2019sentence} as the dense text encoder.

\paragraph{PoT Baseline Implementation.}
Our PoT baseline is implemented following the original prompt design from PoT~\cite{chen2022program}, which provides a general-purpose paradigm applicable to various structured data types. Specifically, the structured data and natural language question are directly provided to the LLM along with a zero-shot prompt that instructs the model to implement a \texttt{``solver()''} function with step-by-step Python program reasoning. The complete prompt template used in our PoT baseline experiments is presented in Table~\ref{tab:pot_prompt}.

\begin{table}[t]
\centering

\footnotesize
\begin{tabular}{p{0.88\columnwidth}}
\toprule
\textbf{PoT Prompt Template} \\
\midrule
\texttt{[data]: \{data\}} \\[4pt]
\texttt{[Question]: \{question\}} \\[4pt]
\texttt{[zero-shot-PoT Prompt]:} \\[2pt]
\texttt{\# Answer this question by implementing} \\[-1pt]
\hspace{0.8em}\texttt{a solver() function.} \\[3pt]
\texttt{def solver():} \\[-1pt]
\hspace{1.5em}\texttt{\# Let's write a Python program step} \\[-1pt]
\hspace{1.5em}\texttt{by step, and then return the answer} \\
\bottomrule
\end{tabular}

\caption{Prompt template for the PoT baseline~\cite{chen2022program}. \texttt{\{data\}} and \texttt{\{question\}} denote placeholders for the structured data and the natural language question, respectively.}
\label{tab:pot_prompt}
\vspace{-1.2\baselineskip}

\end{table}

\begin{figure}[t]
\centering
\includegraphics[width=0.98\columnwidth]{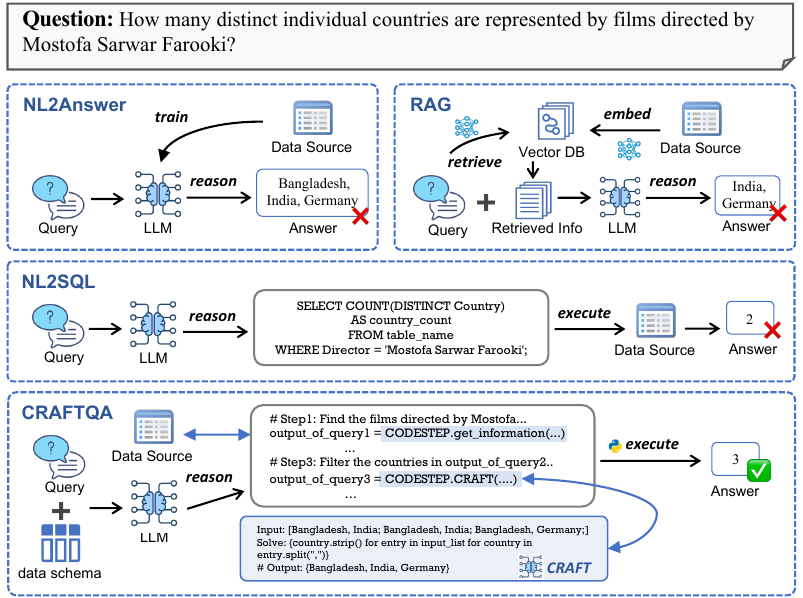} %

    \caption{Comparison of NL2Answer, RAG, NL2SQL paradigm, and our proposed CRAFTQA.}
    \vspace{-\baselineskip}

    \label{fig:intro}
\end{figure}

\begin{table*}[t]
  \centering
  \small
  \setlength{\tabcolsep}{8pt}
  \renewcommand{\arraystretch}{1.2}

  \begin{tabular}{l ccc ccc ccc}
    \toprule
    \multirow{2}{*}{\textbf{Method}} & \multicolumn{3}{c}{\textbf{Fact Checking}} & \multicolumn{3}{c}{\textbf{Num. Reasoning}} & \multicolumn{3}{c}{\textbf{Overall Metric}} \\

    \cmidrule(lr){2-4} \cmidrule(lr){5-7} \cmidrule(lr){8-10}
    
    & \textbf{CDA} & \textbf{DA} & \textbf{F1} & \textbf{CDA} & \textbf{DA} & \textbf{F1} & \textbf{CDA} & \textbf{DA} & \textbf{F1} \\
    \midrule
    
    \multicolumn{10}{l}{\textit{Open-Source Models}} \\
    \midrule
    CRAFTQA-LLaMA-3.1-8B & 25.0 & 57.3 & 61.2 & 30.7 & 31.1 & 32.5 & 29.6 & 36.2 & 38.1 \\
    CRAFTQA-Qwen2.5-7B   & 28.6 & 57.9 & 62.9 & 21.3 & 35.4 & 36.7 & 22.7 & 39.8 & 41.8 \\
    CRAFTQA-Qwen3-32B    & 32.4 & \textbf{72.9} & \textbf{76.2} & 43.4 & \textbf{53.8} & \textbf{54.4} & 41.3 & \textbf{57.5} & \textbf{58.7} \\
    CRAFTQA-DeepSeek-V3  & 30.4 & 70.8 & 71.5 & 36.9 & 48.2 & 48.9 & 35.6 & 52.6 & 53.3 \\

    \midrule
    \multicolumn{10}{l}{\textit{Closed-Source Models}} \\
    \midrule
    CRAFTQA-Qwen-Max         & 61.8 & 75.0 & 76.9 & 51.2 & 48.1 & 49.0 & 53.3 & 53.3 & 54.4 \\
    CRAFTQA-Gemini-2.5-Flash & 48.8 & 75.0 & 76.0 & 49.0 & 54.8 & 55.4 & 49.0 & 58.7 & 59.4 \\
    CRAFTQA-Gemini-2.5-Pro   & 69.7 & 76.0 & 76.6 & 61.7 & 55.1 & 55.3 & 63.3 & 59.2 & 59.5 \\
    CRAFTQA-GPT-3.5          & 18.8 & 61.5 & 65.4 & 33.3 & 38.6 & 40.1 & 30.5 & 43.1 & 45.0 \\
    CRAFTQA-GPT-4o-mini      & 33.3 & 64.6 & 70.2 & 33.0 & 40.7 & 42.6 & 33.1 & 45.4 & 48.0 \\
    CRAFTQA-GPT-4o           & 20.0 & 68.8 & 71.6 & 50.0 & 51.3 & 51.9 & 44.1 & 54.7 & 55.7 \\
    CRAFTQA-o4-mini          & 50.0 & \textbf{83.3} & 83.9 & 48.3 & 56.4 & 56.7 & 48.6 & 61.6 & 62.0 \\
    CRAFTQA-GPT-5-mini       & 45.5 & 82.3 & \textbf{84.9} & 50.0 & \textbf{57.8} & \textbf{59.2} & 49.1 & \textbf{62.6} & \textbf{64.2} \\

    \bottomrule
  \end{tabular}

  \caption{Generalization evaluation of CRAFTQA across different backbone LLMs on TableBench. CDA: Calling Denotation Accuracy, DA: Denotation Accuracy, F1: F1 Score.}
  \label{tab:craftqa_generalization}
\vspace{-1.2\baselineskip}

\end{table*}

\begin{figure}[t]
\centering
\includegraphics[width=0.9\columnwidth]{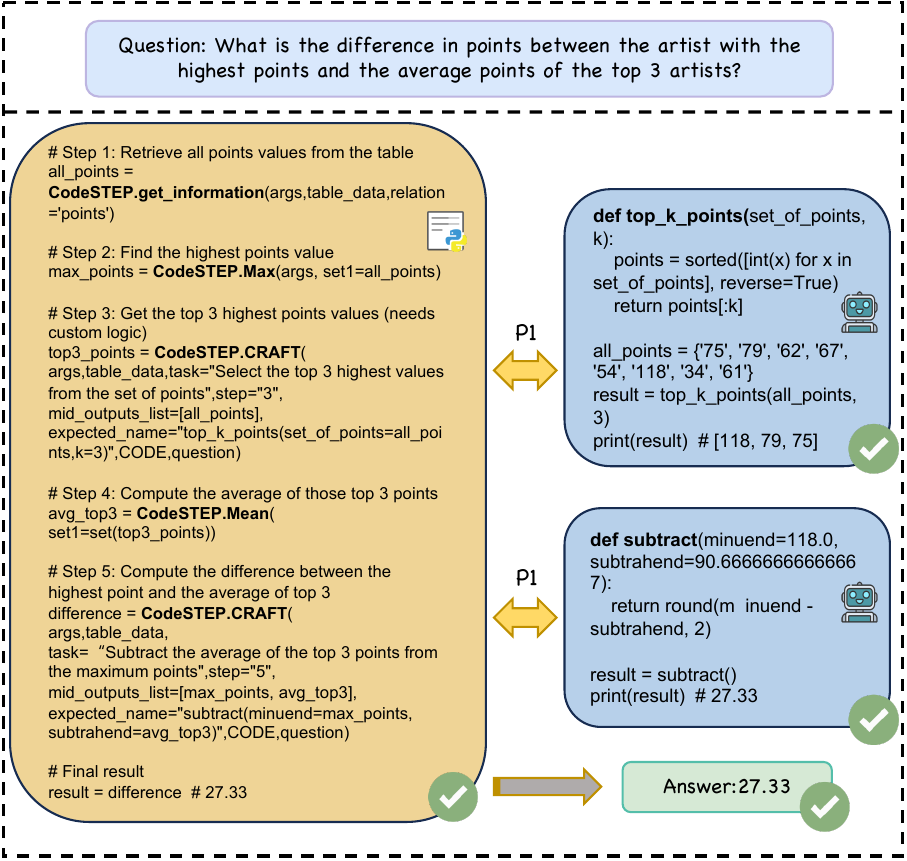} 
    \caption{A more complex positive case study of the CRAFTQA framework}
    \label{fig:case_study_2}
\end{figure}

\section{CDA Overall Assessment}

Table \ref{tab:CRAFTQA_with_TrustUQA_CDA_results} presents the complete numerical results corresponding to Figure \ref{fig:CDA_metrics_comparison_with_sotaTrustUQA} in the main text. This table provides a detailed comparison between CRAFTQA and the current state-of-the-art published method TrustUQA \cite{zhang2025trustuqa} across three datasets (FactChecking, Numerical Reasoning, and WikiSQL-E) under different backbone LLMs (GPT-3.5-Turbo, GPT-4o-mini, and GPT-4.1). The evaluation specifically targets ``out-of-predefined'' scenarios where reasoning operations extend beyond predefined operator sets. Results are presented in the format of \textit{TrustUQA / Ours}, and the $\uparrow$ symbol indicates that CRAFTQA achieves superior performance compared to the baseline. As shown in the table, CRAFTQA consistently outperforms TrustUQA across nearly all metrics and configurations, demonstrating its effectiveness in handling ``out-of-predefined'' reasoning scenarios.



\section{Generalization Across Backbone LLMs}

Table \ref{tab:craftqa_generalization} presents the complete numerical results corresponding to Figure \ref{fig:CRAFT_generalization} in the main text. This table evaluates the generalization capability of CRAFTQA across a diverse range of backbone LLMs, including both open-source models (LLaMA-3.1-8B, Qwen2.5-7B, Qwen3-32B, DeepSeek-V3) and closed-source models (Qwen-Max, Gemini-2.5-Flash/Pro, GPT-3.5, GPT-4o-mini, GPT-4o, o4-mini, GPT-5-mini). The evaluation is conducted on TableBench across Fact Checking and Numerical Reasoning tasks, reporting CDA, DA, and F1 metrics. The results demonstrate that CRAFTQA maintains robust performance across various backbone LLMs with different scales and architectures, and consistently benefits from the advancement of LLM capabilities, confirming its strong generalization potential.

\section{Different Paradigm}

Our method represents a fundamental evolution of the NL2Query paradigm beyond traditional approaches (e.g., NL2SQL \cite{liu2024survey} and NL2SPARQL \cite{jung2020automated}), as illustrated in Figure \ref{fig:intro}.

\section{Case Study}
\label{sec:appendix_case_study}

To further demonstrate CRAFTQA's capability in handling complex multistep reasoning tasks, we present an additional positive case for the question: ``What is the difference in points between the artist with the highest points and the average points of the top 3 artists?'' Figure \ref{fig:case_study_2} illustrates the execution process for this sophisticated query that requires multiple custom operations.

This case exemplifies effective collaboration between CodeSTEP and CRAFT through multiple interactions. CodeSTEP first retrieves all points values from the table and identifies the maximum value using predefined functions. When encountering the need to select the top 3 values, which is a task beyond predefined operations, CodeSTEP appropriately delegates this to CRAFT with clear specifications. CRAFT responds by generating a top ``k'' points function that sorts the points in descending order and returns the first ``k'' elements, successfully extracting ``[118, 79, 75]'' from the data.

After computing the average of these top 3 values using the predefined Mean function, CodeSTEP faces another custom operation requirement: calculating the precise difference between two floating point numbers. Again, recognizing that this exceeds predefined capabilities, CodeSTEP invokes CRAFT with the specific values (``118.0'' and ``90.67''). CRAFT generates a subtract function with appropriate rounding to maintain numerical precision, yielding the final result of ``27.33''.

This case demonstrates several key strengths of the CRAFTQA framework: (1) CodeSTEP's ability to recognize when custom code generation is necessary and provide clear task specifications to CRAFT; (2) CRAFT's capability to generate appropriate functions based on task descriptions and context; (3) The seamless integration of multiple CRAFT invocations within a single reasoning chain; and (4) The framework's effectiveness in handling complex queries requiring both predefined operations and custom implementations. The successful execution showcases how modular design enables flexible problem solving while maintaining computational accuracy throughout the multistep process.

\section{Prompt Template}
\label{sec:appendix_prompt}
\subsection{Prompt for CodeSTEP Generation}  
Table \ref{tab:codestep_prompt} shows the prompt template for CodeSTEP module.

\subsection{Prompt for CRAFT Module}  
Table \ref{tab:craft_prompt} shows the prompt template for CRAFT module.


\begin{table*}[t]
\centering
\scriptsize
\renewcommand{\arraystretch}{1.1}
\setlength{\tabcolsep}{5pt}
\begin{tabular}{p{0.965\textwidth}}
\toprule
\rowcolor{tableheader}\textcolor{white}{\textbf{System Prompt for CodeSTEP Module}} \\
\midrule
\rowcolor{tablesection}\textbf{Role \& Task Description} \\
You are an advanced data analyst proficient in Python, specialized in conditional graph queries for table-based question answering. Your task is to write executable Python code that queries tables to extract relevant information and answer questions based on conditional graph queries. \\
\midrule
\rowcolor{tablesection}\textbf{Core Function Definition} \\
The conditional graph query function is defined as: {\ttfamily CODESTEP.get\_information(args, table\_data=table\_data, relation=None, head\_entity=None, tail\_entity=None, key=None, value=None, tail\_entity\_cmp='=', value\_cmp='=', target\_type=target\_type\_ms, is\_first=False)}. This function retrieves information by querying a data source using the given relation and tail entity as search criteria.
\par\vspace{1pt}
\textbf{Args:}
\newline\hspace{0.8em}{\ttfamily args}: The args parameter is fixed as {\ttfamily args} and cannot be modified.
\newline\hspace{0.8em}{\ttfamily table\_data}: The table\_data parameter is fixed as {\ttfamily table\_data} and cannot be modified.
\newline\hspace{0.8em}{\ttfamily relation} (\textit{str}): The relation to the query that matches the {\ttfamily tail\_entity} or contains the {\ttfamily head\_entity}.
\newline\hspace{0.8em}{\ttfamily tail\_entity} (\textit{str}): The tail entity associated with the relation.
\newline\hspace{0.8em}{\ttfamily head\_entity} (\textit{str}): The head entity that belongs to the relation.
\newline\hspace{0.8em}{\ttfamily key} (\textit{str}): The key to query that matches the {\ttfamily tail\_entity} or {\ttfamily head\_entity}.
\newline\hspace{0.8em}{\ttfamily value} (\textit{str}): The value associated with or belonging to the {\ttfamily key}.
\newline\hspace{0.8em}{\ttfamily tail\_entity\_cmp} (\textit{str}): Comparison operator ({\ttfamily '=', '>', '<', '>=', '<='}), default is {\ttfamily '='}.
\newline\hspace{0.8em}{\ttfamily value\_cmp} (\textit{str}): Comparison operator ({\ttfamily '=', '>', '<', '>=', '<='}), default is {\ttfamily '='}.
\newline\hspace{0.8em}{\ttfamily target\_type}: Fixed as {\ttfamily target\_type\_ms} and cannot be modified.
\newline\hspace{0.8em}{\ttfamily is\_first} (\textit{bool}): Set to {\ttfamily True} for the first query.
\par\vspace{1pt}
\textbf{Returns:} A set of query results.
\par\vspace{1pt}
\textbf{Usage Notes:}
\newline 1) {\ttfamily relation} + {\ttfamily tail\_entity}: `relation' is a column name, and `tail\_entity' is a specific value in that column. This mode returns a set of row identifiers where that column matches the value, e.g.\ {\ttfamily \{`[line\_2]', `[line\_7]', `[line\_1]'\}}.
\newline 2) {\ttfamily relation} + {\ttfamily head\_entity}: `relation' is a column name, and `head\_entity' is one or more row identifier(s) in the {\ttfamily `[line\_id]'} format. This mode returns a set of values from the specified column for those rows. \\
\midrule
\rowcolor{tablesection}\textbf{Constraints \& Usage Guidelines} \\
{[}Note 1{]}: The first call to the {\ttfamily get\_information} function requires {\ttfamily is\_first=True}.
\newline{[}Note 2 - Strict Constraint{]}: In the {\ttfamily get\_information} function, {\ttfamily tail\_entity} and {\ttfamily head\_entity} must never be used together in a single query.
\par\vspace{1pt}
Please follow these guidelines:
\newline 1. Try to use the functions in the provided preset function list to solve the query at each step.
\newline 2. If the preset functions are insufficient, you may use {\ttfamily CODESTEP.CRAFT()} to process the query.
\newline 3. Use Set and Calculator functions as necessary to complete the task.
\newline 4. Record whether {\ttfamily CODESTEP.CRAFT()} was used by setting the ``{\ttfamily use\_CRAFT}'' variable to {\ttfamily True} or {\ttfamily False}. \\
\midrule
\rowcolor{tablesection}\textbf{Preset Function List} \\
\textit{Conditional Graph Query functions:}
\newline $\bullet$~{\ttfamily CODESTEP.get\_information(args, table\_data=table\_data, relation=None, head\_entity=None, tail\_entity=None, key=None, value=None, tail\_entity\_cmp='=', value\_cmp='=', target\_type=target\_type\_ms, is\_first=False)}
\par\vspace{1pt}
\textit{Set functions:}
\newline $\bullet$~{\ttfamily CODESTEP.set\_union(set1, set2, set3=None, set4=None, set5=None)}: Get the union of multiple sets. Returns a set.
\newline $\bullet$~{\ttfamily CODESTEP.set\_intersection(set1, set2, set3=None, set4=None, set5=None)}: Get the intersection of multiple sets. Returns a set.
\newline $\bullet$~{\ttfamily CODESTEP.set\_difference(set1, set2)}: Get the difference between two sets. Returns a set.
\newline $\bullet$~{\ttfamily CODESTEP.set\_negation(table\_data, set1)}: Get all rows except those in {\ttfamily set1}. Returns a set.
\par\vspace{1pt}
\textit{Calculator functions:}
\newline $\bullet$~{\ttfamily CODESTEP.Min(args=args, set1)}: Get the smallest element in {\ttfamily set1}. Returns a set of a numeric value.
\newline $\bullet$~{\ttfamily CODESTEP.Max(args=args, set1)}: Get the largest element in {\ttfamily set1}. Returns a set of a numeric value.
\newline $\bullet$~{\ttfamily CODESTEP.Mean(set1)}: Get the average value of all elements in {\ttfamily set1}. Returns a set of a numeric value.
\newline $\bullet$~{\ttfamily CODESTEP.Count(set1)}: Get the number of elements in {\ttfamily set1}. Returns a set of a numeric value.
\newline $\bullet$~{\ttfamily CODESTEP.Sum(set1)}: Get the sum of elements in {\ttfamily set1}. Returns a set of a numeric value.
\par\vspace{1pt}
\textit{If further assistance is needed, use the} {\ttfamily CODESTEP.CRAFT()} \textit{function:}
\newline $\bullet$~{\ttfamily CODESTEP.CRAFT(args, table\_data=table\_data, task, step, mid\_outputs\_list, expected\_name, CODE\_file\_name=CODE\_file\_name, question\_file\_name=question\_file\_name)}: \newline\hspace{1.8em}\textbf{Args:}
\newline\hspace{3.2em}{\ttfamily args}: Fixed as {\ttfamily args} without modification.
\newline\hspace{3.2em}{\ttfamily task}: The description of the current task step.
\newline\hspace{3.2em}{\ttfamily step}: The current step number.
\newline\hspace{3.2em}{\ttfamily mid\_outputs\_list}: The intermediate results prior to the current step.
\newline\hspace{3.2em}{\ttfamily expected\_name}: Expected function name with parameters.
\newline\hspace{3.2em}{\ttfamily CODE\_file\_name}: Fixed as {\ttfamily CODE\_file\_name}.
\newline\hspace{3.2em}{\ttfamily question\_file\_name}: Fixed as {\ttfamily question\_file\_name}.
\newline\hspace{1.8em}\textbf{Returns:} {\ttfamily mid\_result}: A set or string of results. \\
\midrule
\rowcolor{tablesection}\textbf{Code Generation Guidelines} \\
$\bullet$~Think step-by-step and decompose the problem.
\newline $\bullet$~Only use functions from the provided {[}preset function list{]} to complete the task.
\newline $\bullet$~Use the provided functions to generate Python code directly.
\newline $\bullet$~Only generate Python code; any additional content must be commented with `{\ttfamily \#}'.
\newline $\bullet$~If {\ttfamily CODESTEP.CRAFT()} is used, record it in the `{\ttfamily use\_CRAFT}' variable. \\
\bottomrule
\end{tabular}

\caption{Prompt template for the CodeSTEP module. The system prompt instructs the backbone LLM $\mathcal{M}_\theta$ to generate step-by-step executable Python code for structured data reasoning using predefined functions and the CRAFT interface.}
\label{tab:codestep_prompt}
\vspace{-1.2\baselineskip}

\end{table*}

\begin{table*}[t]
\centering

\scriptsize
\renewcommand{\arraystretch}{1.1}
\setlength{\tabcolsep}{5pt}
\begin{tabular}{p{0.965\textwidth}}
\toprule
\rowcolor{tableheader}\textcolor{white}{\textbf{System Prompt for CRAFT Module}} \\
\midrule
\rowcolor{tablesection}\textbf{\# Task Context} \\
You are an intelligent code generator that produces step-by-step solutions based on multi-stage task descriptions. Focus exclusively on handling the current step task. \\
\midrule
\rowcolor{tablesection}\textbf{\# Input Modules} \\
\textbf{{[}Final Question{]}}
\newline - Represents the final question that needs to be solved in {[}Complete Code{]}.
\newline - It is for reference. You need to provide the processing code and results required for the current step task in {[}Current Task{]}.
\par\vspace{1pt}
\textbf{{[}Complete Code{]}} (Code Framework)
\newline - Complete code representation, so that you can better understand the overall processing logic.
\newline - You need to implement the code representation of the corresponding {\ttfamily CODESTEP.CRAFT()} function in {[}Complete Code{]} and get the result.
\par\vspace{1pt}
\textbf{{[}Current Task{]}}
\newline - The task description that needs to be processed in the current step.
\newline - You need to write the code based on the task description in {[}Current Task{]} with the information and constraints provided.
\par\vspace{1pt}
\textbf{{[}Expected Function Name{]}}
\newline - The function name and parameter representation example expected by this task step, for reference.
\par\vspace{1pt}
\textbf{{[}Previous Steps and Results{]}}
\newline - The output results of each step before the current step.
\newline - The specific functions and code implementation of each step are shown in {[}Complete Code{]}.
\newline - {[}Previous Steps and Results Notes{]}:
\newline\hspace{1em}1. You can get the specific data required for the current step processing from {[}Previous Steps and Results{]}.
\newline\hspace{1em}2. In {[}Previous Steps and Results{]}, if a step result does not contain data of type {\ttfamily `[line\_id]'}, then the original result is directly represented, that is, {\ttfamily `stepx':\{\{`original result of this step'\}\}}.
\newline\hspace{1em}3. In {[}Previous Steps and Results{]}, if a step result contains data of type {\ttfamily `[line\_id]'}, which means the data of a row, then the data information corresponding to each column of the row will be represented accordingly (column name: value), that is, the result of this step will be represented as {\ttfamily `stepx':\{\{`original result of this step':`specific information corresponding to the original result of this step'\}\}}. \\
\midrule
\rowcolor{tablesection}\textbf{\# Output Requirements} \\
\textbf{Code Generation}
\newline 1. Must use {\ttfamily print()} for final result output;
\newline 2. Code must be self-contained (executable independently);
\newline 3. Result format: Single-line text/number.
\par\vspace{1pt}
\textbf{Result Handling}
\newline 1. Return ONLY current step's processed result;
\newline 2. Prohibit intermediate processes/explanations;
\newline 3. Output must match the data type required by the current step (e.g.\ a set, a string, etc.), but must never be a dictionary! \\
\midrule
\rowcolor{tablesection}\textbf{\# Processing Rules} \\
1. Prioritize input data from {[}Previous Steps and Results{]}.
\newline 2. Ensure output directly contributes to solving {[}Final Question{]}.
\newline 3. Solve the current task by writing code based on the information and constraints provided. \\
\bottomrule
\end{tabular}

\caption{Prompt template for the CRAFT module. The system prompt guides the LLM $\mathcal{M}_{\theta'}$ to generate self-contained custom code functions for reasoning steps beyond predefined operations.}
\label{tab:craft_prompt}
\vspace{-1.2\baselineskip}

\end{table*}

\end{document}